\documentclass{article}
\usepackage{ijcai20}

\usepackage{times}

\usepackage{soul}
\usepackage{url}
\usepackage[hidelinks]{hyperref}
\usepackage[utf8]{inputenc}
\usepackage[small]{caption}
\usepackage{graphicx}
\usepackage{amsmath}
\usepackage{booktabs}
\title{Let Me At Least Learn What You Really Like: Dealing With Noisy Humans When Learning Preferences}

\author{
Sriram Gopalakrishnan, Utkarsh Soni
}

\begin{document}

\maketitle

\begin{abstract}
Learning the preferences of a human improves the quality of the interaction with the human. The number of queries available to learn preferences maybe limited especially when interacting with a human, and so active learning is a must. One approach to active learning is to use uncertainty sampling to decide the informativeness of a query. In this paper, we propose a modification to uncertainty sampling which uses the expected output value to help speed up learning of preferences. We compare our approach with the uncertainty sampling baseline, as well as conduct an ablation study to test the validity of each component of our approach.
\end{abstract}

\section{Introduction}

In an AI system that involves a human-in-the-loop, learning the human's preference allows for a better experience for the person. Consider the example of an AI agent for selecting online ads for a user. This is a very noisy environment, and queries with feedback are few and precious. Users seldom want to click on ads much less provide feedback. Advertisements on Facebook for example, do get user feedback. The user has the option to click \textit{"Why am I seeing this ad"}, and provide feedback.

So when the amount of feedback is sparse, rather than trying to learn the entire preference model of the user accurately, if we can at least learn what the important features and weights are, we can better serve the user. Additionally, we should avoid showing product ads that the user hates, and learn a useful model as quickly as possible. 

\begin{figure}[ht]
\centering
     \includegraphics[width=\columnwidth]{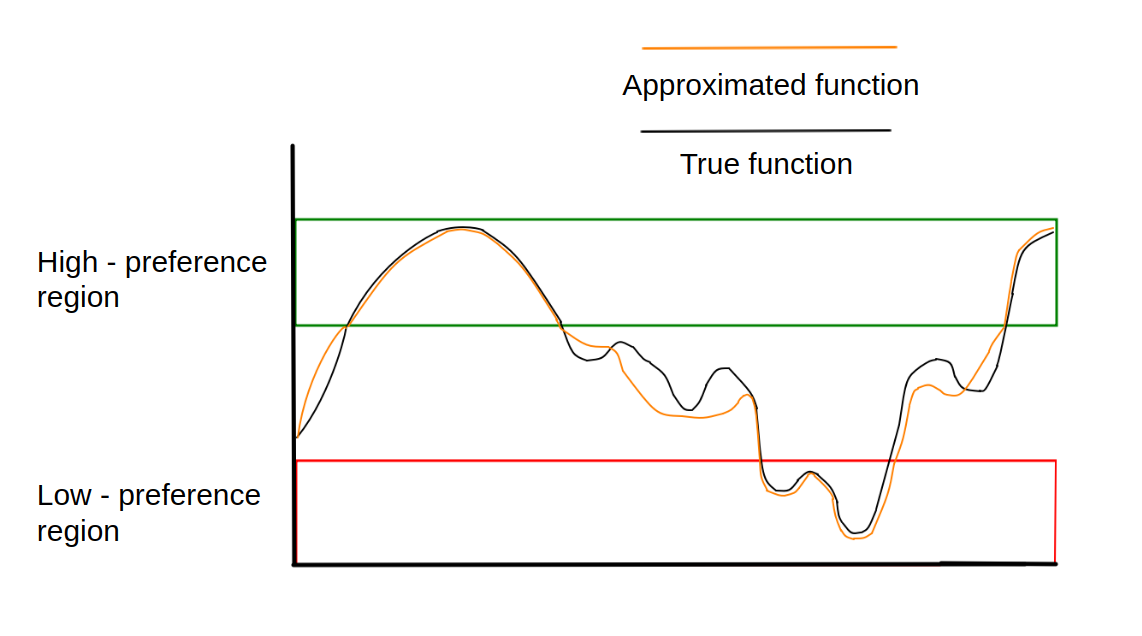}
      \caption{Illustration of Output Value Biased Uncertainty Sampling }
      \label{fig:OBUS_figure.png}
\end{figure}

As to what is a useful model, let's say the human's preference is modelled by a function over some features with noise (as we expect that human feedback is typically noisy). If we had to choose $output$ regions in which to be more accurate, then we argue that it is more important to be accurate when the output (preference) is very high or very low, as compared to being accurate about moderate or low preferences. We justify this with the following arguments. (A) Given a set of options, the best options should be presented, or if there are no clear good options, then bad options should be avoided. (B) Rather than just detecting (classifying) a liked or disliked sample point, we argue that being accurate in the extreme regions of preferences is also important. This is apparent in online ad selection as showing the best ad increases the likelihood of clicking the ad.

We illustrate the idea of a useful function in the figure \ref{fig:OBUS_figure.png} , where the green region is the high preference region, and the red region is the low preference region. We try to learn a better approximation function (the orange line) that matches the true function more closely in those extreme output regions. We tradeoff between accuracy in all regions and sample efficiency by prioritizing accuracy when the output value (magnitude) is higher. In this way we can reduce the number of samples needed to learn a \textit{useful} model of the human's preferences quickly. We leverage this intuition, as well as the idea of uncertainty sampling from Active Learning (AL) to query with the most valuable and informative data points. 

In Active Learning queries are carefully selected so as to have the greatest expected improvement in the accuracy. The queries are posed to an oracle (a source of ground truth information, which could be humans) that gives feedback to help improve the model. It has been widely used to improve the sample efficiency of machine learning models \cite{settles2009active} as well as choosing what experiments to do in the design of experiments. One of the most commonly used AL approaches for sampling queries is uncertainty sampling \cite{lewis1994sequential}. In this approach, the data point sampled is the one for which the model is least confident about it's output value (i.e. has the most variance). In our work we modify uncertainty sampling in a way that is especially applicable to learning preferences. We call our approach Output-Biased Uncertainty Sampling(\textit{OBUS}) . \textit{OBUS} modifies the uncertainty (variance) measure of informativeness by adjusting it based on the output value. In the context of learning preferences, a query is more informative if the variance is higher \textit{and} the \textit{expected} output is a higher value (in magnitude). This deceptively simple change produces valuable gains in learning speed and sample complexity. We will discuss the details further in the methodology section where we also describe the additional details needed for learning preferences with this sampling paradigm.

Note that Active Learning is still a form of supervised learning (as is apparent). A supervised learning task traditionally involves querying only about the label of a data point, and active learning typically does the same. However, there has been several more recent works in active learning like \cite{hema2006active}, \cite{poulis2017learning}, \cite{druck2008learning}, \cite{druck2009active}, \cite{mann2008generalized} where the human annotator is also asked to select relevant(predictive) features for a particular label. In terms of learning preferences, we can liken this to the human explaining their preferences by what features contributed to it. As one might expect, this would greatly speed up learning. There are already both empirical \cite{hema2006active} \cite{druck2009active}, \cite{mann2008generalized} and theoretical works\cite{poulis2017learning} that demonstrate this.We too utilize feature-feedback to speed up the learning.

When the human is queried with a data point, that query is given a preference score or level by the user according to their preference function over some (possibly all) of the features in the plan. The learning problem is then to learn this function. We also enable feature-feedback from the user to tell us what features were liked and disliked. We assume that we start with a large set of features, $F$, that could possibly be relevant (prior knowledge accumulated from past data) of which only a small subset is relevant to the current human's preferences. Then for learning preferences, each data point or query is then just about a subset of features from the set $F$ that are found in the query. 

For example, when a query or sample is presented to the user, there could be many possible features that are present in it. When the query is given to the human, they provide their preference score as well as telling us of salient features that were liked or disliked. 

Let us consider an online ad or message about a local French restaurant. The human could have rated the ad as $4.5$ and indicated that French related ads are liked, and food related ads are disliked. The human does not tell us how much each feature contributed to the score, just that the score reflects their overall feeling about the ad. Note that the range of preference values is not restricted, and it is the relative values of the preferences that define the preference model. The range can be any arbitrary range, but needs to be set prior to any queries. In the online ad scenario it makes sense to restrict this range from 0 to 100. We go into the details in the methodology and experimental sections.

Our claim is that using \textit{OBUS}, we can frugally query the human to quickly learn a useful preference model. We compare our approach with the standard uncertainty (variance) based sampling for the regression problem that we will shortly describe. We also conduct an ablation study to analyze the effect of the different components of our approach. In order to effectively compare the effect of different components and methods of sampling, we use a simulated user. This helps us get copious and comparable feedback for each experimental setting. The feedback from our simulated user is intentionally made noisy in order to mimic a human's noisy feedback. 

Our paper is organized as follows; we first define the problem formally in section $3$ and go over our approach in detail in section $4$. The experimental results are presented in section $5$ where we compare OBUS learning rate to random sampling and traditional uncertainty sampling baselines. We follow this with an ablation study to help evaluate the components of our active learning approach. We then compare our work with existing literature on active learning and eliciting preferences before concluding with a summary, discussion of results, and future directions for this line of work.

\section{Problem Formulation}

In our work the problem of learning preference is represented as a tuple $T = <D,F,O,P,R>$. $D$ is a set of input data points. $F$ is set of features that are present in the data points and possibly relevant to the preferences of the user; $O$ is an oracle that provides feedback (simulated user); $P$ is a probability function that returns the probability of seeing any of an input set of features; The objective is to learn the weights $W$ of a linear preference function over the features in $F$ (not all of which will be relevant to the user/oracle). We will now discuss each of the problem formulation elements in detail.
%P is probability of the UNION of the set of features
% i.e. P(A,B) = P(A) + P(B) - P(A intersection B). 
%Recall the last term is p(A|B)*p(B) In this work we consider the features to be independent, which matches the simulated user.

In our experimental setting, we are given a large pool of data points $D$ split into a training set $D_{train}$ and test set $D_{test}$. The features in each data point come from the superset of features $F$. Some subset of $F$ is relevant to the oracle whose preferences we wish to learn. The features could be arbitrarily complex like compound logical statements that hold true or not, but the complexity of features is an orthogonal dimension that we do \textit{not} explore. In our work, the features are binary features, i.e. the features is either present or not. Our work is not limited to binary features as the user will see from the methodology section; we chose binary features as it is simpler, sufficient, and captures the type of features one might expect in online ads. 

As for the simulated user, which we will henceforth call the \textit{oracle}, it is defined as $O = <p_{relevant},O_{\mu},O_{\sigma},N>$. The oracle is intended to let us evaluate our method against the other methods with the same consistent feedback. The oracle gives feedback on the preference for the queries posed to it. It returns a preference value for a given query data point based on the oracle's value function and also gives feature feedback. The feature feedback would tell us which features were relevant. The oracle is configured with a likelihood $p_{relevant}$ of selecting a feature as relevant. Then the relevant feature is either liked or disliked with equal ($50\%$) chance. The Oracle's preference for an input query is computed as a linear function over the relevant features in the query. Each feature's weight (coefficient) is fixed before the first round of queries, and assigned from a guassian distribution whose properties can be specified by $O_{\mu},O_{\sigma}$. Our reasoning for choosing a gaussian distribution is that it would give us a few features that are high, and low, and most relevant features would have comparable weights. 

Once the magnitude of the feature weight is obtained from the Gaussian distribution, the feature weight is set to be positive if liked and negative if disliked. The preference value returned from the oracle is noisy as human feedback typically is. This preference value is a measure of how strongly the agent likes or dislikes the query. The noise in the preference value is determined by $N$, which represents the standard deviation of a Gaussian noise with mean $0.0$. The oracle samples from this distribution and adds the noise in the preference value feedback. 

We assume that the oracle's preference or value function is linear on the features; $V(x) = W^T \cdot F(x)$, where $W$ represents the weights of the features, and $F(x)$ represents the feature vector of the query $x$ (representing the relevant features). We think using a linear function is an acceptable simplification since features can be arbitrarily complex and non-linearity can be represented as more complex features. This dimension is orthogonal to our contribution.

The objective is to learn a set of weights such that the test set error (in $D_{test}$) is minimized especially when the output preference is an extreme value (very high or very low).In order to prioritize the accuracy in the extreme value regions, we use the value-biased error ($Error_{VB}(.)$). It is computed as the product of the error and true value as described in \ref{eq:value_based_error}.

\begin{equation}
    Error_{VB}(\hat{y},y*) = |\hat{y} - y*| \cdot |y*|
    \label{eq:value_based_error}
\end{equation}

where $\hat{y}$ is the estimated value and $y*$ is the true value. The error for the entire dataset, is the averaged value-biased error from the top $20\%$ and bottom $20\%$ of the range of values. Lastly, we assume the test set follows the same distribution as the training set. 

\section{Methodology}

We discuss the methodology we used by first presenting an overview of the algorithm, and then going into it's parts in detail.

To learn which features matter and their weights, the queries are presented to the Oracle in rounds, in $k=5$  plans per round (hyperparameter). The queries for each round are selected based on the score given by Equation \ref{eq:OBUS_full}. The equation is discussed in the subsection $Scoring Plans$. For now, suffice to say the score represents the informativeness of a query. The score is computed using a model learned by Ridge Regression (which we will call RRM) and using the standard error of the weights to compute the uncertainty in the prediction. We describe this more in the section on Ridge Regression. The oracle gives feedback as to what features were relevant and the score of the plan. After every batch of queries, we train a new RRM model with the additional information. The updated model is used to score the pool of plans. Then the most informative $k$ plans from the remaining are selected for the next batch. 

%Todo
%The general algorithm is described in Algorithm \ref{alg:OBUS_overview}

%Todo
% ADD PSEUDOCODE FORMALIZING THIS DESCRIPTION.
% SAY THE CODE WILL BE PROVIDED UPON PUBLICATION OF THE PAPER

Note that for the first batch of queries, since we have no data, the queries will be selected based on the likelihood of occurrence of their features. The more the probability of occurrence, the more important it is to know about the feature.
Additionally, each round will include a query that is purely exploratory, i.e. tries to cover as many features from the superset $F$ as possible, chosen using the exploration score $S_e$ described in Equation \ref{eq:exploration_score}

\begin{equation}
    S_e(x) = \Sigma_{f \in unseen(x)}p(f)
    \label{eq:exploration_score}
\end{equation}
where $unseen(X)$ are the features in the query $x$ that have not yet been seen by the user (and so we are unsure if they are relevant). This helps find more features to update the RRM model. We now describe the RRM model and variance computation in more detail. 

\subsection{Ridge Regression Model (RRM) }
%feedback and priors
Ridge Regression is appropriate for modelling data of the form in Equation \ref{eq:lin_model_w_noise}
\begin{equation}
    y = w_0 + w \cdot x + \epsilon
    \label{eq:lin_model_w_noise}
\end{equation}
where $w_0$ is the bias which we set as 0 in our experiments, i.e. we assume the human is neither positively or negatively biased. $w$ is the vector of weights for the relevant features in $x$. $\epsilon$ is the noise, assumed to be Gaussian as $N(0,\sigma_{noise})$.  Ridge Regression tries to optimize the loss as shown in  Equation \ref{eq:ridge_regr_loss}

\begin{equation}
    L(X,Y) = \frac{1}{N}\sum_{i=1}^{N} (y_i* - w_0 - w \cdot x_i)^2 + \lambda\sum_j(|w_j|^2)
    \label{eq:ridge_regr_loss}
\end{equation}

which is the mean squared error plus a penalty $\lambda$ on the sum of the squared weights of the model. This penalizes large weights, and implicitly that means we assume that no one weight is extremely large. We used the Ridge Regression implementation in the python library Sklearn\cite{scikit-learn}.
%  This is an often used prior. We do have the option of not using this penalty, and just doing ordinary least squares (OLS) regression, but for our experiments, 
% Additionally, Ridge Regression is adept at dealing with the multicollinearity problem that OLS would struggle with.
If the reader is interested in more details on Ridge Regression, we recommend the lecture notes by Wessel van Wieringen \cite{RidgeRegression}.

Ridge Regression can give us the maximum likelihood estimate with noisy data, where the noise is gaussian with mean zero. 
The confidence range of each parameter is computed using the standard error (se) of each parameter as in Equation \ref{eq:std_err}, which is a well known and often used metric. 
\begin{equation}
    se(w_i) = \sqrt{\frac{\hat{\sigma^2}}{S_{xx}}}
    \label{eq:std_err}
\end{equation}
where $\sigma = \sum_{i=1}^N(y*-y_i)/(N-2)$ is called the residual and an unbiased estimate of the error. $S_{xx} = \sum_{i=1}^N(x_i-\Bar{x})^2$.

For a desired confidence interval defined by $100(1-\alpha)\%$, the range of parameter values is given by the following Equation \ref{eq:param_range}
\begin{equation}
    \hat{w_i} - t_{\alpha/2,n-2}\cdot se(w_i) \leq w_i* \leq \hat{w_i} + t_{\alpha/2,n-2}\cdot se(w_i)
    \label{eq:param_range}
\end{equation}

We used the $90\%$ confidence interval for each parameter using  $\alpha = 0.1$. For each round and for that round, we independently sample parameter values uniformly in this range and stored $N_m=10$ linear models to represent the possible models given the data.Then in order to compute the uncertainty (variance) in output given a data point, we compute the output based on each of the sampled models, and then compute the variance of the outputs. The larger the number of models, the more accurate the variance estimate. For our experiments we chose $N_m=10$ as a tradeoff between speed and efficacy of the estimate for sample selection.
% Additionally, we tried using a Bayesian Linear Model (BLM)\cite{blm_source} as well, but found it too time consuming and computationally expensive. Additionally, the RRM model often matched or outperformed the BLM model, and is much faster to work with.
Initially, the model has no features. As we get feedback on what features are relevant after each round, we add those (as variables) into the model and retrain the model on all the feedback.

\subsection{Scoring Plans}
For sampling by uncertainty, one typically uses the uncertainty (variance) in the prediction as the deciding factor; the more uncertain we are, the more informative that sample is. The main change to the sampling process that we propose and is our core contribution is an Output-Biased Uncertainty Sampling (OBUS). In OBUS, the informativeness of a sample $X$ is computed using 3 scores. The base information score ($S_b$) of a sample $x_i$ is defined in Equation \ref{eq:S_b}.
\begin{equation}
    S_b(x_i) = \sigma(x_i) + \sigma(x_i)^{(\hat{y_i}/\hat{y}_{max})}
    \label{eq:S_b}
\end{equation}
where $\hat{y}_{max}$, is the maximum output value predicted in the pool of queries. So the exponent in the second variance can be at most 1. This limits the contribution of the output value in the base score to at most doubling the standard deviation.
$S_b$ replaces just using variance as the sampling criterion.

The feature frequency score ($S_f$) is the sum of the probabilities of the relevant features in the sample that have been discovered from user feedback. This is shown in Equation \ref{eq:S_f}. This score gives importance to learning about the relevant features that occur more often. 
\begin{equation}
    S_f(X) = \Sigma_{f \in rel(X)}p(f)
    \label{eq:S_f}
\end{equation}
where $rel(X)$ returns the set of relevant features in $X$, and $p(f)$ returns the probability of the feature $f$.
The discovery score ($S_d$) is the sum of probabilities of the features in the sample that have not yet been shown to the user. This is used to explore features that have not yet been shown to the user. $S_d$ is formalized in the Equation \ref{eq:S_d}. Note that this is different from the exploration query described earlier and scored with $S_e$ given in Equation \ref{eq:exploration_score}. Given two queries of equal base score and feature frequency scores, the query with features that have a higher probability mass will get selected. This helps squeeze as much information value from a query as possible.
\begin{equation}
    S_d(X) = \Sigma_{f \in unseen(X)}p(f)
    \label{eq:S_d}
\end{equation}
where $unseen(X)$ returns the set of features in $X$ which have been not yet been shown to the user.

The total information score ($S_t$) of a data sample $X$ is the base score adjusted with the feature frequency score and discovery score, as shown in Equation \ref{eq:OBUS_full}.

\begin{equation}
    S_t(X) = S_b(X)*(1 + S_f(X) + S_d(X))
    \label{eq:OBUS_full}
\end{equation}

This $S_t$ score is what is used to rank and select the queries from the data pool $D$ for the next round. We will make our code available to the public after publication of this work

\section{Experiments and Discussion}

In our experiments we compare our method of $OBUS$ that uses the informativeness score defined in Equation \ref{eq:OBUS_full} against two baselines; sampling by uncertainty (just variance) and random sampling. We also conduct an ablation study to analyze the effect of the two modification scores, i.e. discovery score $S_d$ and feature frequency score $S_f$

Please note that when we compare with uncertainty sampling method, we include the feature discovery score $S_d$ and feature frequency score $S_f$ to make it a fair comparison, without which we get an unfair advantage. In fact the informativeness score used in the uncertainty sampling method is the same as in Equation \ref{eq:OBUS_full} with the difference being in the base score $S_b$. The base score is set to just the variance which represents the uncertainty.

For our experiments, there are $|F| = 200$ features. Our pool of queries is of size $|D|=10,000$ and each query contains 4 features. We set the probability of a feature being selected by the oracle as relevant is $p_{relevant} = 0.1$ (10\% of all features). The mean and variance of the Oracle's feature weights is set as $O_\mu = 8.0$,$O_\sigma = 3.0$. The Oracle's noise in ratings is intentionally set high to $N = 6.0$ to test the methodology in the extreme. We run each of our experiments for $30$ trials and present the averaged data.

% Finally, we also perform an ablation study to show the effect of each of the components in the informativeness score.

\subsection{Comparing $OBUS$, with Uncertainty Sampling and Random Sampling}

\begin{figure}[!ht]
\centering
     \includegraphics[width=0.6\columnwidth]{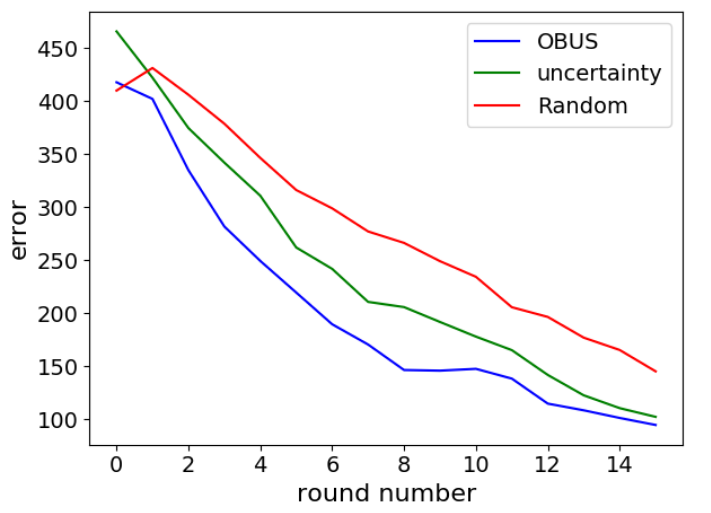}
      \caption{Comparison of OBUS with Baselines using Value-Biased Error ($Error_{VB}$) }
      \label{fig:baselin_comp_region}
\end{figure}

\begin{figure}[!ht]
\centering
     \includegraphics[width=0.6\columnwidth]{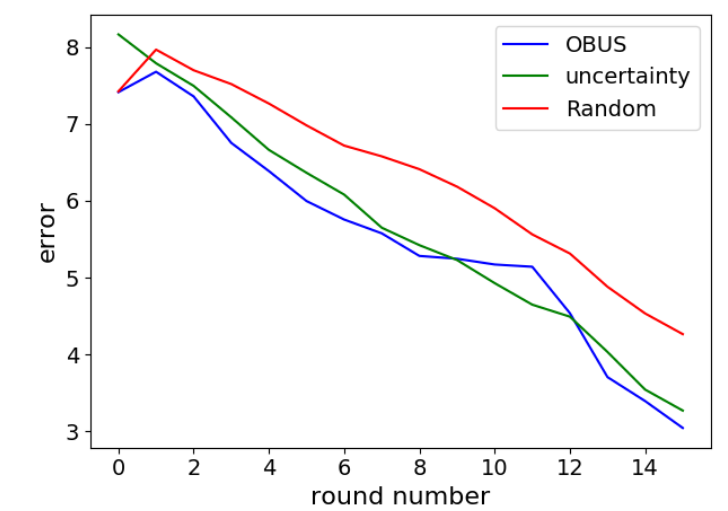}
      \caption{Comparison of OBUS with Baselines using Average Error in All Regions }
      \label{fig:baselin_comp_total}
\end{figure}
% \begin{figure}[!ht]
% \centering
%      \includegraphics[width=\columnwidth]{scenario_2.png}
%       \caption{Scenario-2; a subset of features are relevant }
%       \label{fig:with_irr}
% \end{figure}
Using the aforementioned settings, we display the results of the comparison with the baselines in Figure \ref{fig:baselin_comp_region} and Figure \ref{fig:baselin_comp_total}. As the reader can see from Figure \ref{fig:baselin_comp_region}, \textit{OBUS} offers a modest but real benefit over vanilla uncertainty sampling in the extreme regions, which is measured using $Error_{VB}$ as defined in Equation \ref{eq:value_based_error}. The range of values may be confusing to the reader; please note that the output values in our experiments lie (on average) in the range [-28,28]. The untrained model's prediction is zero, and so $Error_{VB}$ would be $28^2 = 784$ and the model's $Error_{VB}$ would drop from there after the first round. The important point to note is the relative difference between \textit{OBUS} and other baseline methods, which is consistent across all rounds. This data is after averaging over $30$ trials. This demonstrates an increased accuracy in the extreme output regions. The improvements are modest improvements, and this is while staying comparable to uncertainty sampling in the overall error as Seen in Figure \ref{fig:baselin_comp_total}. In the context of our running example of online ad-selection, these modest improvements in the extreme region could translate to a small increase in the odds of a user clicking on an add. Over a large user base, this translates to a lot more clicks and revenue for the host website.

Random sampling does quite well too, as one might expect. Random sampling discovers features well, and samples the frequent features more often. It is still noticeably worse than the other sampling methods. We observed that as the size of the feature set $F$ increases and the number of relevant features decreases, random sampling suffers more as one might expect. 

\subsection{Ablation Study}
\begin{figure}[!ht]
\centering
     \includegraphics[width=0.60\columnwidth]{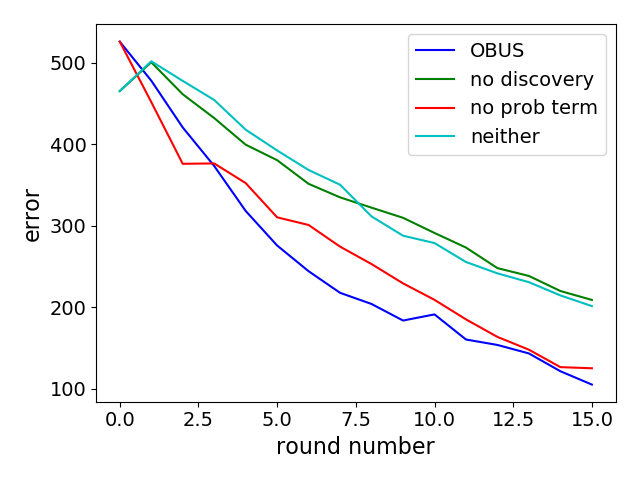}
      \caption{Ablation Study Over the Terms of OBUS Total Score using Value-Biased Error ($Error_{VB}$) }
      \label{fig:ablation_graph_region}
\end{figure}

\begin{figure}[!ht]
\centering
     \includegraphics[width=0.60\columnwidth]{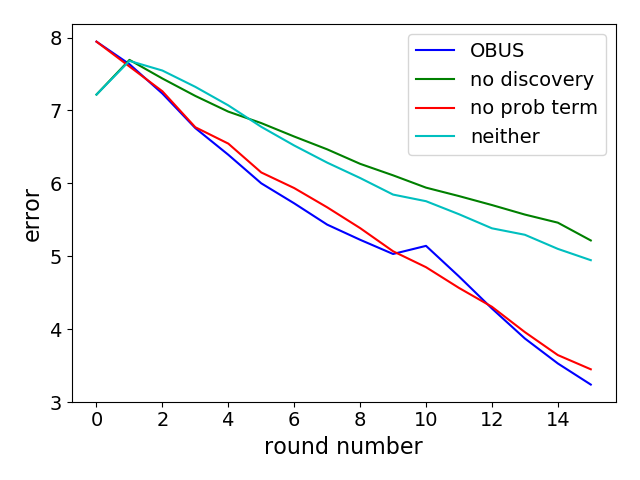}
      \caption{Ablation Study Over the Terms of OBUS Total Score using Average Error in All Regions}
      \label{fig:ablation_graph_total}
\end{figure}

The total score $S_t$ used in \textit{OBUS} (Equation \ref{eq:OBUS_full}) is comprised of two additional terms $S_f$, and $S_d$. We analyze the effect of these terms in the OBUS score function by comparing the performance of OBUS removing one or both of these terms. The results are shown in Figure \ref{fig:ablation_graph_region} and Figure \ref{fig:ablation_graph_total}. The second term is the known-feature occurrence term which prioritizes learning about relevant features that occur more often. The third term is the feature discovery term which helps in feature discovery by giving weight to those samples that have frequent features that the user has not specified as relevant or not. To arrive at these results, we follow the same process of taking the average error for each round over 30 trials, as was done in the previous experiment. 

As evidenced in the Figure \ref{fig:ablation_graph_region},the full \textit{OBUS} score ($S_t$) performs better than all in the extreme regions based on $Error_{VB}$. It is closely followed by the trendline \textit{"no prob term"}, which corresponds to the \textit{OBUS} score that turns off the term with $S_f$. The \textit{OBUS} score that uses feature frequency score $S_f$ (\textit{"no discovery"}) but drops discovery score $S_d$ performs worse because it did not discover as many relevant features. This is the same for \textit{OBUS} with only the base score (cyan trendline titled \textit{"neither"})

In terms of overall error (displayed in Figure \ref{fig:ablation_graph_total}), the full \textit{OBUS} score $S_t$ it is comparable to the method that only keeps $S_d$ (discovery) score. Both discover as many relevant features, and the accuracy overall is about even. As expected, without discovering these features, the other two cases suffer.

It must be noted that $S_f$ score without $S_d$ does not confer much benefit, but combined together, they give the total \textit{OBUS} score $S_t$ a benefit in the extreme regions as in Figure \ref{fig:ablation_graph_region}.

\section{Related Work}
In this section, we focus on discussing the relevant Active Learning literature with respect to preference learning, using feature feedback, and Active Learning for regression.

Active learning techniques are often used to accelerate preference learning (of humans) as we expect human feedback to be scarce. For instance, \cite{maystre2017just} proposed an efficient method for ranking a list of items in the \textit{noisy} setting where a pairwise comparison query may be answered incorrectly. They achieved improvements in the sampling efficiency as compared to random sampling. However, our preference setting is starkly different and arguably more general as we are learning a preference function, not just a ranking over a set of points. 

A different approach or dimension in Active Learning is using feature-feedback to improve the learning rate. In the approach proposed by \cite{hema2006active} for document classification, the learner interleaves the feature query and the instance query. Instance query ask for the label of a document, whereas feature query asks about the relevance of features (a feature is relevant if it is highly indicative for the label). To incorporate the feature-feedback, the relevant feature's values are scaled for all the data points. This boosts the accuracy of the document classifier. The way we use feedback in our work has parallels to what they do. In our approach we obtain feedback on features that affect the user's preferences, and use the feedback to eliminate irrelevant features from consideration in subsequent query selection. Additionally, their approach is suitable for classification task while our method is meant for a regression task. In learning preferences, it is important to know the order of preference for the options and not just if it is preferred or not; so their method would not work well.  

Another work that leverages feature feedback for active learning is that of \cite{mann2008generalized}. In it, the authors use labelled features to train a conditional random field for a sequence labelling task. A conditional distribution of labels given a feature is learnt as a reference distribution. Feature-feedback is then incorporated into the learning model by trying to match this distribution to the conditional distribution of labels over the unlabelled data. This learning paradigm gives substantial improvement in sample efficiency. They show that the same accuracy as traditional instance labelling can be achieved with $10$ times fewer annotations used. A similar approach was used in \cite{druck2008learning} for a classification task. It reports as high as $20$ times speed-up in time when using labelled features over labelled instances. They assume that labeling a feature is 5 times faster than labeling a document (a result supported empirically by \cite{hema2006active}). 

The task in the previous three approaches is labelling and classification, and their approaches cannot be applied to learning a preference order over data. However, the biggest difference is in the way we do uncertainty sampling where we utilize the expected $output$ values and not just variance (uncertainty). This makes our approach suitable for learning preference models where there is a benefit of being more accurate in the extreme output regions, as in the case of online ad selection; the AI system only needs to select one/few highly preferred ads.   %Another point of difference is that these works use the expected model difference to select samples. This is fundamentally different from uncertainty sampling. The last difference is in how they use feature feedback; they use it to just identify indicator or relevant variables, whereas we use feature feedback in a way that is specific to learning preferences; to determine if the feature is liked or disliked. This gives an informative bias to our prior, and helps in learning a Bayesian Linear Model. The feedback process is tailored to learning preferences.

Given these works, we can say that the literature on AL has convincingly shown that feature feedback is essential in speeding up the learning rate. Moreover, some theoretical bounds have also been provided for some settings like document classification \cite{poulis2017learning}. 

% In addition to other works done which empirically show the benefit of using feature-feedback \cite{druck2009active, settles2011closing} for active learning, the work in \cite{poulis2017learning} studies the concept of feature-feedback theoretically. It formalizes two approaches for feature feedback in a document classification setting: one based on probabilistic disjunction model(PDM) and other using a SVM. For the PDM model, a theoretical bound was given for the number of annotations needed by the user to learn the true label for a topic(the approach essentially converts document to a topic distribution which are used to predict the label). For the SVM with feature-feedback, they give formal proof of reduction in the bound of the loss. 

A more closely related work which combines the regression problem setting and feature feedback is that of  \cite{craig}. In their work, they learn a function to recommend products that are represented by features and concepts. A concept is a subjective property like "safety of the car" which can depend on values of different features for different users. The utility of a product is given by the weighted sum of features and a bonus for satisfying the concept. Given the uncertainty in weight values, and which features are relevant to the concept(referred to as versions of concepts), the approach recommends a product which minimizes the maximum regret(MMR). The maximum regret is the maximum possible difference between utilities of $x^*$(currently recommended product) and some adversarially chosen  $x^a$(chosen over all possible weights and concept versions). They propose several query strategies to efficiently minimize the regret value of the recommended product $x^*$. This is very different from the uncertainty sampling variation that we use. Additionally their objective is only to recommend the most preferred product. They only elicit positive preference information. We learn a model with both positive and negative features, and seek to be accurate in predicting both high $and$ low preference cases. 

The most important distinguishing feature of our approach overall is that our sampling is sensitive to the output region in which we expect the input to lie. We found one other work that considers sampling based on the output region by Eric et al. \cite{nando_GP}. In that work, the authors use Gaussian Processes (GP) to search for a single region of high preference. They focus on the expected output value to guide the learning, which is similar to us. They are different in the sense that the search is for a single setting of parameters, rather than learning weights on the parameters (which are features for us). Our approach lets us identify many highly preferred points as opposed to one parameter setting. They explicitly acknowledge that they do not focus on learning the valuation function, since their problem has an infinite parameter space. Ours has a discrete and finite parameter space which makes the problem easier but and still useful. Additionally, their use of Gaussian Processes makes it suitable \emph{only} when the number of parameter/features are few and manageable (as stated in their work) because GP are computationally very demanding. 

To the best of our knowledge , our intuitive modification to uncertainty sampling has not been explored in the literature, and it contributes to learning preferences faster for the extreme regions using the methodology we described.

\section{Conclusion and Future Work}

In summary, our $OBUS$ approach is a method of active learning for preferences that becomes more accurate in the high and low output regions faster than using uncertainty sampling. \textit{OBUS} modifies uncertainty sampling by including the expected output value term as an adjustment. We also consider the probability of feature occurrence, and the discovery value of queries to select queries. We empirically demonstrated that our method works well as compared to the standard uncertainty sampling baseline and random sampling. We demonstrated consistent results using simulated user feedback. We also conduct an ablation study to show the validity of the additional terms for feature discovery and feature frequency. 

%negative bias in online ads. So negative bias should be on. Doesnt affect the sampling process, and in our experiments it is true, but need not be

%adding weights in the base score. Normalizing the gain differently.

With respect to the running example of online ad selection, the web service can learn a personal model of preferences using a methodology like ours. This personal model can be used on top of other recommendation algorithms that the service uses, which would leverage demographic information,history and location. Such heuristic information can be used to display exploratory ads, before building confidence in the user's model and using it instead.

An additional benefit of our approach is that the uncertainty in the output can also be used to choose what options are presented. If the situation is higher risk/cost, then a more certain option that is highly preferred can be given; when the cost of failure is lower, then a more uncertain option with high expected value can be given. 

Our work can be extended in terms of the feedback interface, by supporting pairwise comparisons or using discrete levels of preference instead of asking the user to return a number between a fixed interval like [0,100]. This would make the feedback process better for the user. 

% \begin{acks}
%   Zero acks given
% \end{acks}

\bibliographystyle{named}
\bibliography{ijcai20.bib}

\end{document}